\documentclass[journal]{IEEEtran}
\usepackage{amsmath,amsfonts}
\usepackage{algorithmic}
\usepackage{algorithm}
\usepackage{array}
\usepackage[caption=false,font=normalsize,labelfont=sf,textfont=sf]{subfig}
\usepackage{textcomp}
\usepackage{stfloats}
\usepackage{url}
\usepackage{verbatim}
\usepackage{graphicx}
\usepackage{setspace}
\usepackage[numbers]{natbib}
\usepackage{multirow}
\usepackage{amssymb} 
\usepackage{hyperref}
\usepackage{color}
\usepackage{xspace}
\usepackage{svg}
\begin{document}

\title{AdvLogo: Adversarial Patch Attack against Object Detectors based on Diffusion Models}

\author{
    \IEEEauthorblockN{Boming Miao\IEEEauthorrefmark{1}, Chunxiao Li\IEEEauthorrefmark{1}, Yao Zhu\IEEEauthorrefmark{2}*, 
    Weixiang Sun\IEEEauthorrefmark{3}, Zizhe Wang \IEEEauthorrefmark{2}, Xiaoyi Wang\IEEEauthorrefmark{1}, Chuanlong Xie\IEEEauthorrefmark{1}} \\
    \IEEEauthorblockA{\IEEEauthorrefmark{1}Beijing Normal University} \\
    \IEEEauthorblockA{\IEEEauthorrefmark{2}Tsinghua University} \\
    \IEEEauthorblockA{\IEEEauthorrefmark{3}Northeastern University}
}

\maketitle

\begin{abstract}
With the rapid development of deep learning, object detectors have demonstrated impressive performance; however, vulnerabilities still exist in certain scenarios. Current research exploring the vulnerabilities using adversarial patches often struggles to balance the trade-off between attack effectiveness and visual quality. To address this problem, we propose a novel framework of patch attack from semantic perspective, which we refer to as AdvLogo. Based on the hypothesis that every semantic space contains an adversarial subspace where images can cause detectors to fail in recognizing objects, we leverage the semantic understanding of the diffusion denoising process and drive the process to adversarial subareas by perturbing the latent and unconditional embeddings at the last timestep. To mitigate the distribution shift that exposes a negative impact on image quality, we apply perturbation to the latent in frequency domain with the Fourier Transform. Experimental results demonstrate that AdvLogo achieves strong attack performance while maintaining high visual quality.
\end{abstract}


\section{Introduction}
\IEEEPARstart{I}{n} recent years, deep neural networks have significantly advanced the field of computer vision \cite{krizhevsky2012imagenet,he2016deep}. As network-based models become increasingly integrated into daily life, concerns about privacy have grown \cite{shokri2015privacy,fredrikson2015model}, leading to the proposal of adversarial attack to protect data from being over-exploited by neural networks \cite{szegedy2013intriguing}. Adding imperceptible noise to images is an effective way to fool classifiers \cite{goodfellow2014explaining}. However, this approach is not suitable for object detectors that capture images in real-time. Adversarial patches (AdvPatch) represent a classical technique for evading detection by object detectors and have demonstrated practical applications in physical attacks \cite{thys2019fooling,huang2023t,xu2020adversarial}. Despite their effectiveness, AdvPatch often exhibits disorganized and conspicuous patterns, increasing the risk of being segmented \cite{liu2022segment}. To enhance the imperceptibility of AdvPatch, \citet{hu2021naturalistic} introduce NAP, which utilizes GANs to generate adversarial patches that incorporate semantic information. Nonetheless, there is still considerable room for improvement in the attack performance of NAP.

\begin{figure}[!t]
    \centering
    \includegraphics[width=0.99\linewidth]{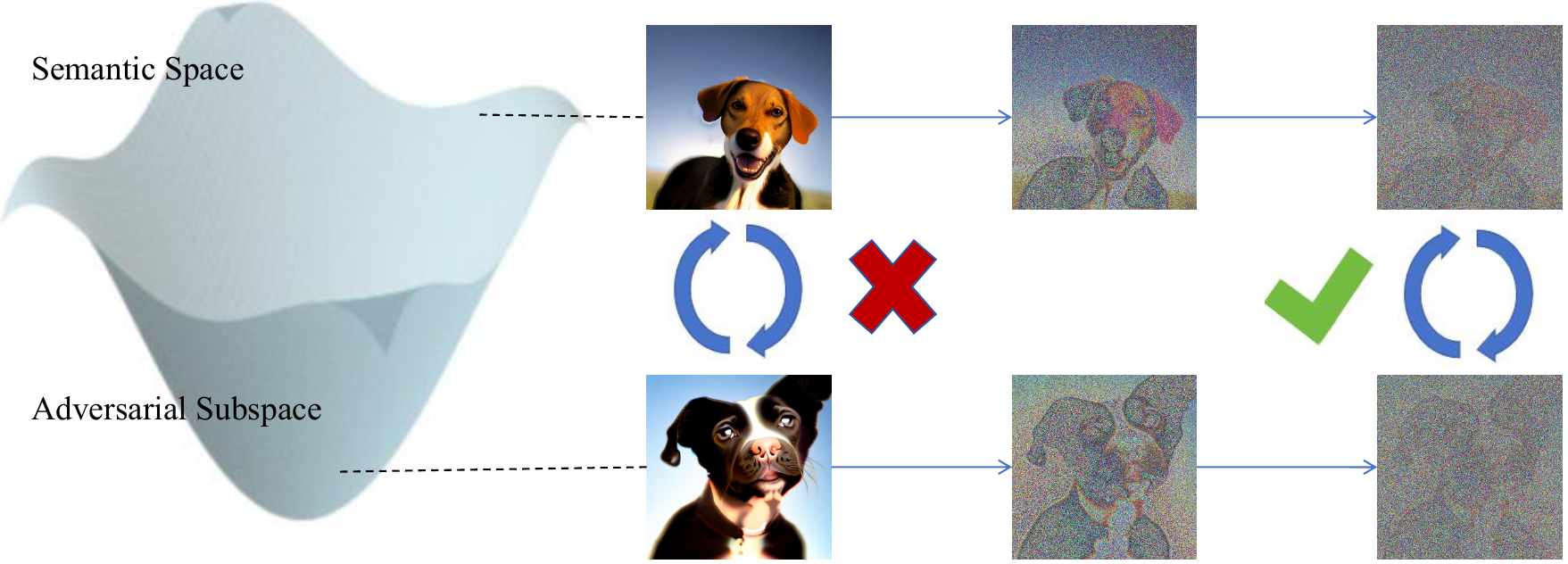}
    \caption{According to our hypothesis, the semantic space of "dog" contains an adversarial subspace. The visual appearances of two dogs from different regions of this space are distinct, making it impossible to transfer from one to the other under normal conditions. However, when significant noise is introduced, the distinctions between the noisy versions of the two dogs become less pronounced, making the transfer between these regions possible.}
    \label{fig:motivate}
\end{figure}

In this work, we propose AdvLogo to balance the trade-off between attack performance and visual quality. We hypothesize that every semantic space contains a subspace that is adversarial to models. Specifically, as shown in Fig.~\ref{fig:motivate}, we suggest that not every image of a "dog" can be correctly handled by the models. There is a subspace within which the image may be recognized as a dog by humans, yet it could interfere with the object detectors, causing them to misclassify or fail to recognize the object. When significant Gaussian noise is added to images, the differences between these images diminish, and they gradually approach a Gaussian distribution, facilitating the transfer between images. Therefore, we propose a method of transferring normal images to adversarial ones within the noisy image space, followed by generating clean adversarial samples with specific semantic information using the denoising process of diffusion models. Classic diffusion models use latent variables, unconditional embeddings, and conditional embeddings to guide the denoising process, where conditional embeddings are crucial for maintaining the correctness of the semantic space; hence, we keep them fixed. Our approach minimizes the confidence of the target object detector as the optimization objective, employing gradient information to optimize the latent variables in the frequency domain. This strategy aims to enhance the adversarial effectiveness of the samples while minimizing the impact on the latent distribution, thus preserving image quality. Additionally, we find that synchronizing this process with the unconditional embeddings optimization significantly boosts the adversarial effectiveness while maintaining visual quality. Given the high computational cost and memory requirement of gradient calculation during the denoising process of diffusion models, we derive a simple gradient approximation method based on the chain rule, suitable for updating both the latent variables and unconditional embeddings. Our contributions are summarized as follows:
\begin{enumerate}
\item We introduce a novel patch attack framework against object detectors from a semantic perspective. By optimizing the noise in frequency domain, we significantly enhance the visual quality of adversarial patches while achieving effective adversarial transferability.
\item We thoroughly analyze the impact of optimizing the unconditional embeddings on the adversarial effectiveness of images. Optimizing the unconditional embeddings can significantly enhance its adversarial capability, with minimal degradation in the aesthetic quality of AdvLogo.
\item Extensive experiments demonstrate that AdvLogo consistently outperforms NAP in terms of attack efficacy, and achieves competitive attack performance compared to AdvPatch while offering enhanced visual fidelity and superior robustness against defense mechanisms.
\end{enumerate}

The rest of the paper is organized as follows. In Section~\ref{sec:related work}, we introduce the details of related work. And then we introduce our method in Section~\ref{sec:method}. The main experimental results are presented in Section~\ref{sec:experiment}. We briefly discuss the prospect of future work in Section~\ref{sec:future work}. Finally, we present our conclusions in Section~\ref{sec:conclusion}.

\section{Related Work}
\label{sec:related work}
\subsection{Transfer-based Attack}
Black-box attack can be categorized into transfer-based and query-based. Since attackers cannot fully access the models in real world, transfer-based attack is more practical compared to the latter. In transfer-based attack, a surrogate model is used to craft adversarial examples. However, these examples face the risk of overfitting to the surrogate model during the training stage \cite{papernot2017practical}. Therefore, improving the transferability of adversarial examples has become a key issue, and various perturbation strategies have been proposed to address this challenge \cite{qi2023transaudio, su2019one, sun2024medical}. Many works have explored ways to improve transferability at levels of images and models. \citet{xie2019improving} propose data augmentation by randomly selecting transformations of input data. \citet{tramer2017ensemble} point out ensembling models is also an effective way. In face of limited resource of models, \citet{huang2023t} propose a self-ensemble strategy inspired by shake-drop \cite{yamada2019shakedrop}. From the perspective of gradients, \citet{wu2020skip} observe that utilizing more gradients from skip connections can yield higher transferability. However, these methods remain constrained within the pixel space. Some studies consider exploring the perturbation space. For instance, \citet{chen2024content} demonstrate that perturbations in the content space offer higher transferability compared to the pixel space. \citet{duan2021advdrop} introduce adversarial examples by removing high-frequency information, showing that perturbations in frequency domain are more resistant to JPEG compression \cite{wallace1991jpeg}. \citet{zhu2021rethinking, zhu2022toward} analyze the distribution differences between normal examples and adversarial examples, suggesting that driving data to distributions where models are less well-trained can achieve higher transferability. Furthermore, recent studies \cite{chen2023diffusion, chen2024content, xue2024diffusion} have demonstrated that adversarial attacks on classifiers based on diffusion models can achieve greater imperceptibility and transferability. 

\subsection{Object Detectors}
Object detection has achieved significant advancements with the aid of neural networks \cite{zou2023object}, and these models can be categorized into two types: one-stage and two-stage. One-stage models simultaneously predict the positions and identities of objects, offering a direct and efficient approach to detection. This integration of localization and classification in a single step allows for faster processing, making one-stage models suitable for real-time applications. In contrast, two-stage models first identify potential target areas through region proposal networks and then classify the objects within those areas. This segmented approach often yields higher accuracy and precision, albeit at the cost of increased computational complexity and slower inference times. The first network-based model is R-CNN \cite{girshick2014rich}, marking the inception of the deep learning era in object detection. RCNN initially generates a set of object proposals, each of which is scaled to a fixed size, and subsequently employ linear SVM classifiers to detect the presence of objects within each region and to classify the object categories. \citet{he2015spatial} propose SPPNet, enabling CNN models to generate a fixed-length representation from inputs of varying sizes. Faster RCNN \cite{ren2015faster} improves upon RCNN and SPPNet by introducing a Region Proposal Network (RPN), which allows for nearly cost-free region proposals. FPN \cite{lin2017feature} develops a top-down architecture with lateral connections, building high-level semantics at all scales. In contrast, one-stage detectors are developed to simultaneously retrieve all objects along with their locations and categories. \citet{redmon2016you} propose YOLO, which partitions the image into regions and simultaneously predicts bounding boxes and probabilities for each region. Subsequent versions of YOLO \cite{redmon2017yolo9000,redmon2018yolov3,bochkovskiy2020yolov4} address localization accuracy issues. YOLO v5 \cite{jocher2020ultralytics} further improves the efficiency and speed of YOLO and has been widely applied. \citet{liu2016ssd} propose Single Shot MultiBox Detector (SSD), which introduces detecting features at different layers and achieves multi-reference and multi-resolution detection.

\subsection{Patch Attack}
Adversarial patch attack is a localized perturbation method widely employed in vision-based tasks, particularly due to its potential for extension to physical attack. The approach is first proposed by \cite{brown2017aurko} to attack classifiers. A patch of visible size is scaled and applied to a restricted region of images, and the optimization of the patch can be expressed as
\begin{equation}
    \tau=\underset{\tau}{\arg\max} L\left(f_\theta\left(A(x,\tau)\right)\right)
\end{equation}
where $f_\theta$ is the surrogate model, $\tau$ is the adversarial patch applied to image $x$ with rendering function $A$, and $L$ denotes the loss function. In classification tasks, $L$ is usually set as Cross Entropy Loss \cite{krizhevsky2012imagenet}. \citet{brown2017aurko} first put forward patch attack in classification and demonstrate its effectiveness. However, object detectors aim to identify and localize multiple objects within an image, making the attack surface more complex. Several studies have explored the application of adversarial patches in attacking object detectors. \citet{xie2017adversarial} first propose patch attack in the field of object detection. The same idea is also carried out in \cite{liu2018dpatch}. \citet{thys2019fooling,chen2019shapeshifter} propose methods to apply physical patches for this purpose. These studies demonstrate that carefully designed patches could indeed fool object detectors in practical scenarios. \citet{huang2023t} further enhance the transferability of adversarial patches by incorporating self-ensemble techniques. \citet{wu2020dpattack,huang2021rpattack} compress the perturbed region and perturbed fewer pixels to improve the imperceptibility. However, optimization in pixel space makes these patches vulnerable to detection by patch detectors \cite{liu2022segment}. Once detected and segmented from the image, the effectiveness of the attack is compromised. Given that patches containing meaningful information are less likely to attract attention, they can effectively evade potential defenses. \citet{hu2021naturalistic} propose NAP, generating adversarial patches by optimizing the latent space of GAN, yielding high imperceptibility and robustness but sacrificing the attack performance.

\section{Method}
\label{sec:method}
In this section we give the problem definition in \ref{subsec:preliminary}. Then we introduce our overall frame work in \ref{subsec:advlogo}. The details of optimization are in \ref{subsec:latent}. 
\subsection{Preliminary}
\label{subsec:preliminary}
In the diffusion forward process, an original latent $z_0$ is progressively noised over $T$ steps using a scheduler $\left\{\beta_t: \beta_t \in(0,1)\right\}_{t=1}^T$ through $T$ steps. Let $\alpha_t=\prod_{s=1}^t \left(1-\beta_s\right)$, then
$
z_T = \sqrt{\alpha_T}z_0+\sqrt{1-\alpha_T}\epsilon,
$
where $\epsilon \sim \mathcal{N}(0, \mathbf{I})$.
For sufficiently large $T$, $z_T$ converges to a fixed point given the noise sampled at each timestep, regardless of the distribution of $z_0$. Our approach guides the prediction of the conditional distribution during the denoising process $\Omega$ to create adversarial examples. The conditional distribution depends on the noise predicted by the network $\epsilon_\theta$:
\begin{equation}
    \epsilon_\theta(z_t,C,\Phi_t)=\epsilon_\theta(z_t,\Phi_t) + \omega(\epsilon_\theta(z_t,C)-\epsilon_\theta(z_t,\Phi_t))
\end{equation}
where $\omega$ is the guidance scale, $C$ is the fixed text embedding ensuring consistent semantic information, and $\Phi_t$ is the unconditional embedding at each timestep. For simplicity, we use $\epsilon_\theta(z_t)$ to represent $\epsilon_\theta(z_t,C,\Phi_t)$. Modifying $z_t$ and $\Phi_t$ can generate examples in adversarial subspaces. However, optimization at every timestep incurs high computational cost, so we simplify the process by modifying only the last timestep $T$:
\begin{equation}
z_T^*, \Phi_T^* =\underset{z_T,\Phi_T}{\arg \min }L_{det}\left(f_\theta,x, \Omega(z_T,C,\Phi_T,\ldots, \Phi_1)\right)
\end{equation}
where $L_{det}$ is the loss function measuring the detector's corruption, and $f_\theta$ and $x$ represent the object detector and input images respectively.

\subsection{Overall Framework}
\label{subsec:advlogo}
\begin{figure*}[!t]
\centering
\includegraphics[width=0.98\textwidth]{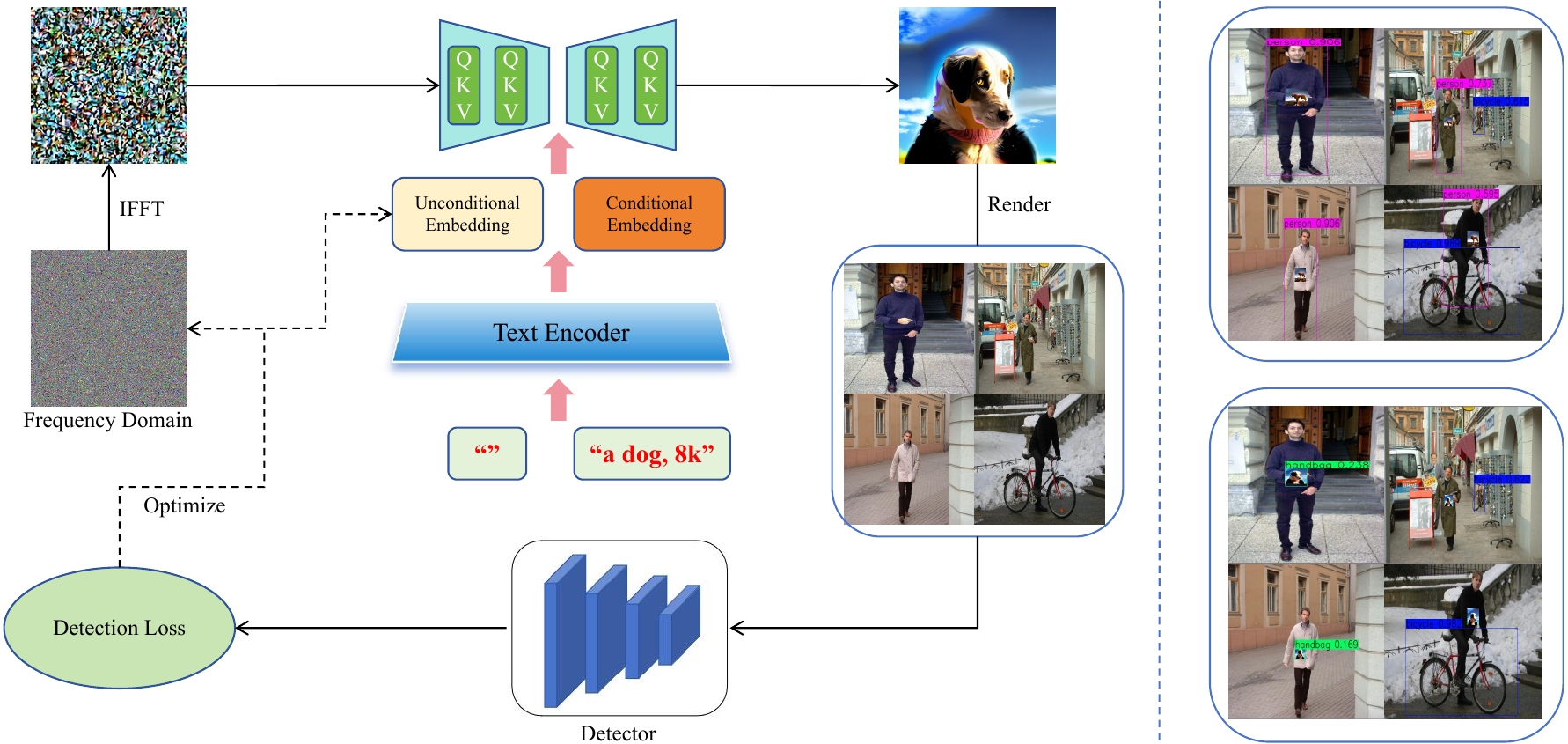}
\caption{During the training stage, we initially obtain a noisy latent representation. This representation is transformed into the frequency domain using the Fourier Transform. Starting from this frequency domain, we retrieve the latent variable $z_T$ by applying the Inverse Fourier Transform. Subsequently, we apply a denoising process to generate the corresponding patch. This patch is rendered onto the target object to create adversarial images. These images are then evaluated by an object detector, yielding a detection loss. The frequency domain representation and unconditional embedding are iteratively optimized to minimize this loss. During the attack stage, when the optimized patch is applied to the target (e.g., a person), object detectors fail to recognize the presence of the target. As a contrast experiment, we apply a non-adversarial patch to the same location, which results in successful detection by the object detectors. This comparison underscores the effectiveness of our adversarial patch at evading detection.}
\label{fig:framework}
\end{figure*}

As described in Fig~\ref{fig:framework}, the training stage begins by generating a noisy latent representation. This noisy latent is then transformed into the frequency domain using the Fourier Transform. From the frequency domain, we reconstruct the latent variable $z_T$
through the Inverse Fourier Transform. This reconstructed latent serves as the basis for generating the adversarial patch via a denoising process. The generated patch is subsequently rendered onto the target object to simulate an adversarial scenario. These adversarial images, with the patch attached, are passed through object detectors, and the resulting detection loss is computed:
\begin{equation}
\label{eq:detection loss}
    L_{det}=\frac{1}{\sum_{i=1}^N M_i}\sum_{i=1}^N\sum_{j=1}^{M_i} conf_{i,j}^2
\end{equation}
where $N$ denotes the number of input images, $M_i$ is the number of detection boxes on the $i$-th image, and $conf_{i,j}$ represents the confidence score of the $j$-th detection box on the $i$-th image. This loss serves as the feedback mechanism for optimizing both the frequency domain representation and the unconditional embeddings. By iteratively minimizing this detection loss, the patch is refined to maximize its adversarial effectiveness, ensuring that the final patch is capable of evading black-box object detectors.

\subsection{Latent Frequency Domain}
Although diffusion models offer opportunities to generate adversarial patches with rich semantic information, they require that the latent variable $z_T$ adheres strictly to Gaussian distribution. Directly perturbing the spatial domain of the latent variable can lead to significant distribution shifts, which can negatively impact the visual quality of the adversarial patches. To mitigate this issue, we consider applying perturbations in the frequency domain, which is less sensitive to numerical variations at adjacent spatial points. To be specific, we first transform the latent variable $z_T$ into the frequency domain $\tilde{z}_T$ using the Fourier Transform. For a 4-dimensional tensor, the transform can be expressed as:
\begin{equation}
\begin{aligned}
&\tilde{z}_T(k_1, k_2, k_3, k_4) = \sum_{n_1=1}^{d_1} \sum_{n_2=1}^{d_2}\sum_{n_3=1}^{d_3}\sum_{n_4=1}^{d_4} z_T(n_1, n_2, n_3, n_4) \\
&\cdot \exp\left(-2\pi i \left(\frac{k_1 n_1}{d_1} + \frac{k_2 n_2}{d_2} + \frac{k_3 n_3}{d_3} + \frac{k_4 n_4}{d_4}\right)\right) \\
&= \sum_{n_1=1}^{d_1} \sum_{n_2=1}^{d_2}\sum_{n_3=1}^{d_3}\sum_{n_4=1}^{d_4} z_T(n_1, n_2, n_3, n_4) \\
&\cdot \left[\cos\left(2\pi \left(\frac{k_1 n_1}{d_1} + \frac{k_2 n_2}{d_2} + \frac{k_3 n_3}{d_3} + \frac{k_4 n_4}{d_4}\right)\right) \right. \\
 &\left. - i\sin\left(2\pi \left(\frac{k_1 n_1}{d_1} + \frac{k_2 n_2}{d_2} + \frac{k_3 n_3}{d_3} + \frac{k_4 n_4}{d_4}\right)\right)\right] \\
 &= \tilde{z}_T^R(k_1, k_2, k_3, k_4)+\tilde{z}_T^I(k_1, k_2, k_3, k_4).
\end{aligned}
\label{eq:fft}
\end{equation}
We can see that every point of $\tilde{z}_T$ contains information from all points of $z_T$, so perturbing $\tilde{z}_T$ has a reduced impact on the distribution of $z_T$ compared to perturbing $z_T$ directly. We adopt the first-order optimization method PGD \cite{madry2017towards} to apply the perturbation:
\begin{equation}
 \tilde{z}_T \gets \tilde{z}_T-\alpha\cdot\textbf{sign}(\nabla_{\tilde{z}_T}L_{det}),
 \end{equation}
 where $\alpha$ is the step size. To prevent significant distribution shifts, we use the $L_\infty$ norm ball as a bound for perturbation. Given an original variable $z$ and a maximum norm $\epsilon$, the clamp operation to ensure that the perturbed variable $y$ remains within the bound can be expressed as::
\begin{equation}
\mathcal{B}_{\infty}(z, \epsilon)=\max (\min (y, z+\epsilon), z-\epsilon).
\end{equation}
As shown in (\ref{eq:fft}), the frequency domain variable $\tilde{z}_T$ can be decomposed into its real component $\tilde{z}_T^R$ and imaginary component $\tilde{z}_T^I$, so we can set bounds for $\tilde{z}_T^R$ and $\tilde{z}_T^I$ respectively. After perturbed within their respective bounds, $\tilde{z}_T$ is then transformed back to the spatial domain using the Inverse Fourier Transform: 
\begin{equation}
\label{eq:ifft}
\begin{aligned}
&z_T(n_1, n_2, n_3, n_4) = \frac{1}{d_1 d_2 d_3 d_4} \sum_{k_1=1}^{d_1} \sum_{k_2=1}^{d_2} \sum_{k_3=1}^{d_3} \sum_{k_4=1}^{d_4} \\
&\tilde{z_T}(k_1, k_2, k_3, k_4) \\ 
&\cdot \exp\left(2\pi i \left(\frac{k_1 n_1}{d_1} + \frac{k_2 n_2}{d_2} + \frac{k_3 n_3}{d_3} + \frac{k_4 n_4}{d_4}\right)\right).
\end{aligned}
\end{equation}
This approach allows us to introduce perturbations that better preserve the original distribution and reduce the negative impacts on the generated images while still achieving the desired adversarial effects.
\subsection{Unconditional Embeddings Optimization}
In addition to $z_T$, which defines the initial state in the Gaussian space during the denoising process, unconditional embeddings play a crucial role in determining the subareas of the generated logos. Thus optimizing these embeddings can help exploit the target model's vulnerabilities more effectively. However, optimizing the embeddings at every timestep would result in significant computational complexity. To simplify the process, we focus on perturbing the unconditional embeddings specifically at the last timestep $z_T$. This approach allows us to achieve the desired adversarial effect without adding unnecessary complexity in the earlier stages. The embeddings $\Phi_T$ are updated iteratively using gradient information as follows:
\begin{equation}
    \Phi_T \gets \Phi_T -\beta\nabla_{\Phi_T}L_{det},
\end{equation}
where $\beta$ is the step size for updating $\Phi_T$. Through iterative optimization, we progressively refine the unconditinal embeddings to maximize their effectiveness in generating adversarial logos.
\subsection{Gradient Approximation}
\label{subsec:latent}
We adopt the Denoising Diffusion Implicit Model (DDIM) \cite{song2020denoising} for the denoising process. Admittedly, the inversion is differentiable, and the gradient $\nabla_{\tilde{z}_T}L_{det}$ and $\nabla_{\Phi_T}L_{det}$ can be derived by the chain rule:
\begin{equation}
\begin{aligned}
    \nabla_{\tilde{z}_T}L_{det}&=\frac{\partial L_{det}}{\partial\tau}\cdot\frac{\partial\tau}{\partial z_0}\cdot\frac{\partial z_0}{\partial z_1}\ldots\frac{\partial z_{T-1}}{\partial z_T}\frac{\partial z_T}{\partial \tilde{z}_T}, \\
    \nabla_{\Phi_T}L_{det}&=\frac{\partial L_{det}}{\partial \tau}\cdot\frac{\partial\tau}{\partial z_0}\cdot\frac{\partial z_0}{\partial z_1}\ldots\frac{\partial z_{T-1}}{\partial \Phi_T}.
\end{aligned}
\end{equation}
However, the computational cost will be huge if we include the entire calculation graph. To address this problem, we simplify the gradient computation with constant approximation. DDIM estimates the sample at the previous timestep using the following equation:
\begin{equation}
\begin{aligned}
    z_{t-1}&=\sqrt{\frac{\alpha_{t-1}}{\alpha_t}}\left(z_t-\sqrt{1-\alpha_t} \epsilon_\theta\left(z_t\right)\right)\\ 
    &+\sqrt{1-\alpha_{t-1}} \epsilon_\theta\left(z_t\right) \\
    &=\sqrt{\frac{\alpha_{t-1}}{\alpha_t}}z_t+\sqrt{\frac{\alpha_t-\alpha_{t-1}}{\alpha_t}}\epsilon_{\theta}\left(z_t\right).
\end{aligned}
\end{equation}
We can get the derivative of the latent of the previous timestep:
\begin{equation}
    \frac{\partial z_{t-1}}{\partial z_t} =  \sqrt{\frac{\alpha_{t-1}}{\alpha_t}}+\sqrt{\frac{\alpha_t-\alpha_{t-1}}{\alpha_t}}\frac{\partial \epsilon_{\theta}\left(z_t\right)}{\partial z_t}.
\end{equation}
For $\frac{\partial z_0}{z_1}\ldots\frac{\partial z_{T-2}}{\partial z_{T-1}}$, we can obtain that
\begin{equation}
\begin{aligned}
    &\frac{\partial z_0}{\partial z_{T-1}}=\left(\sqrt{\frac{\alpha_0}{\alpha_1}} + \sqrt{\frac{\alpha_1-\alpha_0}{\alpha_1}} \frac{\partial \epsilon_\theta\left(z_1\right)}{\partial z_1}\right)\frac{\partial z_1}{\partial z_2}\ldots\frac{\partial z_{T-2}}{\partial z_{T-1}} \\
    &=\left(\sqrt{\frac{\alpha_0}{\alpha_1}}\frac{\partial z_1}{\partial z_2} + \sqrt{\frac{\alpha_1-\alpha_0}{\alpha_1}} \frac{\partial \epsilon_\theta\left(z_1\right)}{\partial z_2}\right)\frac{\partial z_2}{\partial z_3}\ldots\frac{\partial z_{T-2}}{\partial z_{T-1}} \\
    &=\left(\sqrt{\frac{\alpha_0}{\alpha_2}} +\sqrt{\frac{\alpha_2-\alpha1}{\alpha_2}}\frac{\partial \epsilon_{\theta}(z_2)}{\partial z_2} + \sqrt{\frac{\alpha_1-\alpha_0}{\alpha_1}}\frac{\partial \epsilon_\theta(z_1)}{\partial z_2}\right)  \\
    & \frac{\partial z_2}{\partial z_3}\ldots\frac{\partial z_{T-2}}{\partial z_{T-1}} =\sqrt{\frac{1}{\alpha_{T-1}}}+\sum_{i=1}^{T-1}\sqrt{\frac{\alpha_i-\alpha_{i-1}}{\alpha_{i-1}\alpha_i}}\frac{\partial \epsilon_\theta(z_i)}{\partial z_{T-1}},
\end{aligned}
\end{equation}
where $\alpha_0=1$. We can observe that the computational cost is contributed by $\frac{\partial \epsilon_\theta(z_i)}{\partial z_{T-1}}, t=1,2,\ldots, T-1.$ If $\epsilon_\theta(z_i), i=1,2,\ldots T-1$ are seen as constant, these components will be equal to zero. In this way we can take
$
    \frac{\partial z_0}{\partial z_{T-1}}=\sqrt{\frac{1}{\alpha_{T-1}}}, 
$
thus the gradient calculation cam be simplified as:
\begin{equation}
\label{eq:gradient}
\begin{aligned}
    &\nabla_{\tilde{z}_T}L_{det}=\nabla_{z_0}L_{det}\left(\sqrt{\frac{1}{\alpha_{T-1}}} \frac{\partial z_T}{\partial \tilde{z}_T} \right.\\
    &\left. +\sqrt{\frac{\alpha_T-\alpha_{T-1}}{\alpha_{T-1}\alpha_T}}\frac{\partial \epsilon_\theta(z_T)}{\partial \tilde{z}_T}\right) \\
    &\nabla_{\Phi_T}L_{det}=\nabla_{z_0}L_{det}\sqrt{\frac{\alpha_T-\alpha_{T-1}}{\alpha_{T-1}\alpha_T}}\frac{\partial \epsilon_\theta(z_T)}{\partial \Phi_T}
\end{aligned}
\end{equation}
The simplification relies on the condition that the predicted noise remains constant in subsequent steps. However, the condition is not satisfied after perturbations are applied. Consequently, it is necessary to update the noise predicted after each time of perturbation. The simplified DDIM denoising process $\tilde{\Omega}$ can be written as:  
\begin{equation}
\label{eq:ddim_sim}
    \begin{aligned}
        z_{T-1}&=\sqrt{\frac{\alpha_{T-1}}{\alpha_T}}\left(z_T-\sqrt{1-\alpha_T} \epsilon_\theta\left(z_T\right)\right), \\
        &+\sqrt{1-\alpha_{T-1}} \epsilon_\theta\left(z_T\right)\\
        \epsilon_t &\gets \epsilon_\theta(z_t), t=T-1,\ldots, 1.\\
         z_{t-1}&=\sqrt{\frac{\alpha_{t-1}}{\alpha_t}}\left(z_t-\sqrt{1-\alpha_t} \epsilon_t\right) +\sqrt{1-\alpha_{t-1}} \epsilon_t.
    \end{aligned}
\end{equation}
The overall steps for AdvLogo are detailed in Algorithm~\ref{alg:ala}.
\begin{algorithm}
\caption{Adversarial Logo Attack}
\label{alg:ala}
\begin{algorithmic}[1]
\REQUIRE A prompt $P$, a detector $f_\theta(\cdot)$, input data $x$, batch size $bs$, patch applier $A$, denoising step $T$, initial latent at the last timestep $z_T$, attack iterations $K$, max attack epoch $M$, learning rate of the latent $\alpha$, learning rate of the unconditional embeddings $\beta$, perturbation bound $\delta$.
\ENSURE The adversarial logo
\STATE $\tilde{z}_T=FFT(z_T)$ based on (\ref{eq:fft})
\STATE $z:=\tilde{z}_T$
\STATE $z_T=IFFT(\tilde{z}_T)$ based on (\ref{eq:ifft})
\STATE $C,\Phi_t \gets \textbf{TextEncoder}(P,``"), t=1,2,\ldots, T.$
\STATE $\tau \gets \tilde{\Omega}(z_T,C,\Phi_T,\ldots, \Phi_1)$ based on (\ref{eq:ddim_sim})
\STATE $X_1,\ldots,X_{\frac{N}{bs}}\gets \textbf{batch}(x)$
\FOR{$m=1, \ldots, M$}
    \FOR{each $X \in [X_1, \ldots, X_{\frac{N}{bs}}]$}
        \FOR{$k=1,\ldots, K$}
        \STATE $bbox^{clean}, conf^{clean} \gets f_\theta(X)$
        \STATE $X^{adv} \gets A(X, \tau,bbox^{clean})$
        \STATE $bbox^{adv}, conf^{adv} \gets f_\theta(X^{adv})$
        \STATE Calculate loss $L_{det}$ based on (\ref{eq:detection loss})
        \STATE Calculate $\nabla_{\tilde{z}_T}L_{det}$ and $\nabla_{\Phi_T}L_{det}$ based on (\ref{eq:gradient}).
        \STATE $\tilde{z}_T \gets \tilde{z}_T - \alpha \cdot \textbf{sign}\left(\nabla_{\tilde{z}_T} L_{\text{det}}\right)$
        \STATE $\tilde{z}_T^R \gets \mathcal{B}_\infty\left(z^R,\delta\right)$, $\tilde{z}_T^I \gets \mathcal{B}_\infty\left(z^I,\delta\right)$.
        \STATE $z_T=IFFT(\tilde{z}_T)$
        \STATE $\Phi_T \gets \Phi_T - \beta \nabla_{\Phi_T}L_{det}$ 
        \STATE $\tau \gets \tilde{\Omega}(z_T,C,\Phi_T,\ldots, \Phi_1)$
        \ENDFOR
    \ENDFOR
\ENDFOR
\STATE \textbf{return} $\tau$
\end{algorithmic}
\end{algorithm}

\section{Experiment}
\label{sec:experiment}
\subsection{Implementation Details}
\label{subsec:implementation details}
\textbf{Dataset and Models} In accordance with the existing work \cite{thys2019fooling,hu2021naturalistic}, we use the INRIA Person dataset \cite{dalal2005histograms} for our experiments, which comprises 614 images in the training set and 288 images in the test set. Stable Diffusion 2.1 \cite{Rombach_2022_CVPR} is used as the base model in the DDIM denoising process. To assess the effectiveness of our proposed framework, we evaluate it using seven mainstream object detection models, representing both one-stage and two-stage architectures to ensure a thorough evaluation. The models include YOLO v2 \cite{redmon2017yolo9000}, YOLO v3 \cite{redmon2018yolov3}, YOLO v4 \& YOLO v4-tiny \cite{bochkovskiy2020yolov4}, YOLO v5 \cite{jocher2020ultralytics}, Faster R-CNN \cite{ren2015faster}, and SSD \cite{liu2016ssd}. These models encompass a diverse range of techniques and strategies in object detection, including speed-oriented one-stage models and accuracy-focused two-stage models. We evaluate our framework across diverse architectures to showcase its robustness and effectiveness in improving attack performance.

\textbf{Experimental Setting} The target class to attack in our experiments is "person". The size of AdvLogo is 512 $\times$ 512. The training scale of AdvLogo is set as 0.15, and the test scale is 0.2. We set $K=5$, $M=20$, $T=20$, $\alpha=0.2$, $\beta=10^{-4}$ and $\delta=30$. 

\textbf{Evaluation Metrics} We use the mean Average Precision (mAP) to assess the attack capability of the crafted patches. Since black-box attacks are more valuable in the real world, we calculate mAP on models other than the surrogate model. To better demonstrate the impact of the patches on detectors, we consider the detectors’ predictions on clean data as the ground truth (i.e., mAP = 1).

\subsection{Main Results}
\label{subsec:main results}

\begin{table*}
\centering
\caption{Performance Comparison of Different Methods. We generate adversarial examples based on seven distinct models and evaluate our patch across all of these models, calculating the black-box average performance on the remaining six models. We compare the performance of AdvLogo against AdvPatch and NAP on YOLO v2, YOLO v3, YOLO v4, YOLO v4-tiny, YOLO v5, Faster R-CNN and SSD.}
\scalebox{0.82}{ 
\begin{tabular}{c|c|c c c c c c c|c}
\hline
\textbf{Model} &
\textbf{Method} & YOLO v2 & YOLO v3 & YOLO v4 & YOLO v4tiny & YOLO v5 & Faster R-CNN & SSD & \textbf{Black-Box Avg$\downarrow$} \\
\hline
\multirow{3}{*}{YOLO v2} & AdvPatch & 5.66 & 40.26 & 48.49 & 24.44 & 43.38 & 39.27 & 41.28 & \textbf{39.52} \\ 
& NAP & 38.05 & 56.85 & 67.72 & 67.58 & 66.86 & 56.60 & 56.62 & 62.04 \\
& AdvLogo (Ours) & 37.60 & 48.46 & 45.05 & 33.30 & 52.65 & 37.57 & 47.77 & 44.13 \\
\hline
\multirow{3}{*}{YOLO v3} & AdvPatch & 51.85 & 13.89 & 57.16 & 58.43 & 70.47 & 51.46 & 59.79 & 58.19 \\ 
&NAP & 49.87 & 55.39 & 66.96 & 67.06 & 58.69 & 55.35 & 51.18 & 58.19 \\
&AdvLogo (Ours) & 38.05 & 43.01	& 46.11 & 43.29	& 58.25 & 43.05	& 52.75 & \textbf{46.92} \\
\hline
\multirow{3}{*}{YOLO v4} & AdvPatch & 37.41 & 37.18 & 19.67 & 26.91 & 46.37 & 44.67 & 43.07 & \textbf{39.27} \\ 
& NAP & 58.47 & 66.09 & 72.53 & 71.12 & 75.76 & 57.71 & 60.91 & 65.01 \\
& AdvLogo (Ours) & 43.56 & 51.88 & 33.08 & 46.00 & 56.32	& 42.62	& 56.71	& 49.52 \\
\hline
\multirow{3}{*}{YOLO v4tiny} & AdvPatch & 47.41 & 59.59  & 66.48 & 14.50 & 69.51 & 55.95 & 55.22 & 59.03 \\ 
& NAP & 39.48 & 50.28 & 61.93 & 57.58 & 54.47 & 54.62 & 41.15 & 50.32 \\
& AdvLogo (Ours) & 41.93 & 48.04 & 44.40 & 22.51 & 40.34 & 41.80 & 50.58 & \textbf{44.52} \\
\hline
\multirow{3}{*}{YOLO v5} & AdvPatch & 46.90 & 54.93 & 62.16 & 46.15 & 13.39 & 47.71 & 50.73 & 51.43 \\
& NAP &40.55 & 65.39 & 68.61 & 65.85 & 70.10 & 55.83 & 59.42 & 59.28 \\
& AdvLogo (Ours) & 46.66 & 40.85  & 39.47 & 30.95 & 18.66 & 41.17 & 45.46 & \textbf{40.76} \\
\hline
\multirow{3}{*}{Faster R-CNN} & AdvPatch & 24.53 & 23.37 & 26.54 & 25.60 & 30.46 & 8.62 & 42.15 & \textbf{28.78} \\ 
& NAP & 43.62 & 61.56 & 59.28 & 62.68 & 50.80 & 44.13 & 54.71 & 55.44 \\
& AdvLogo (Ours) & 40.60	& 60.63	 & 60.34 & 52.70	& 59.69	& 38.93 & 55.60 & 54.93 \\
\hline
\multirow{3}{*}{SSD} & AdvPatch & 32.43 & 62.93 & 66.22 & 49.07 & 51.19 & 47.42 & 15.10 & \textbf{51.54} \\ 
& NAP & 57.46 & 70.14 & 70.94 & 77.79 & 78.40 & 61.38 & 63.81 & 69.35 \\
& AdvLogo (Ours) & 43.10 & 58.95 &  66.02 & 51.32 & 62.99 & 44.18 & 48.02 & 54.43 \\
\hline
\end{tabular}}
\label{tab:performance}
\end{table*}

We craft our adversarial logos using the prompt "a dog, 8k" based on seven different models. The logo is then placed at the center of the detection boxes generated by these models when detecting objects in clean images. We subsequently evaluate the models' detection performance on images with these adversarial logos, calculating the mAP for each model. For each individual model, the other six models are treated as black-box models, and the average mAP of these six models is used to assess the transferability of the attack performance. As shown in Table~\ref{tab:performance}, compared to NAP, AdvLogo demonstrates superior performance in terms of average black-box mAP, indicating that AdvLogo exhibits better attack effectiveness. The adversarial logos generated by YOLO v2 achieve an average black-box attack performance improvement of 17.91\% over NAP, while those generated by YOLO v4 outperform NAP by 15.49\%. Comparing AdvLogo to AdvPatch, these two methods exhibit similar attack performance. Specifically, the adversarial logos generated by YOLO v3 achieve an average black-box attack performance that is 11.27\% better than AdvPatch, while those generated by YOLO v2 are 4.61\% less effective than AdvPatch in average black-box attack performance. The overall results reveal that AdvLogo demonstrates competitive attack performance compared to AdvPatch, while appearing more natural and less noticeable. In Section~\ref{subsec:robustness}, we demonstrate that our attack method is more robust than AdvPatch.

\begin{figure}[ht]
\centering
\includegraphics[width=0.48\textwidth]{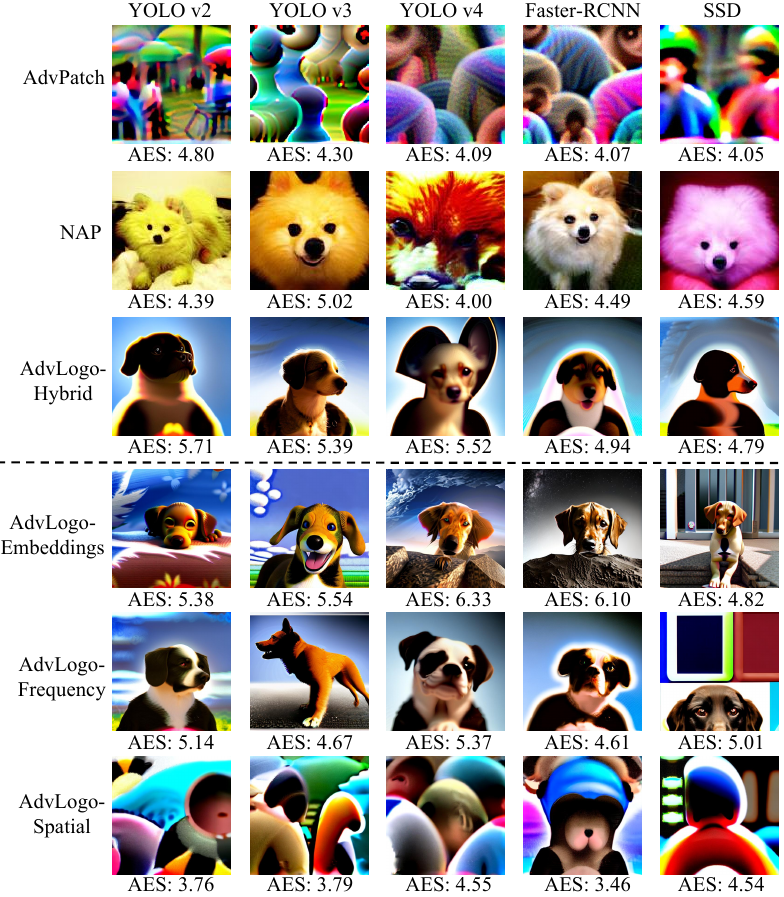}
\caption{Comparison of visual quality of different adversarial patches. Our proposed method is AdvLogo-Hybrid, demonstrates high visual quality. For the ablation study, we also present logos generated with different optimization strategies. Specifically, AdvLogo-Embedding is obtained by optimizing $\Phi_T$ alone, AdvLogo-Frequency is obtained by optimizing $\tilde{z}_T$ alone, and AdvLogo-Spatial is obtained by optimizing $z_T$ alone. Both NAP and AdvLogo-Hybrid exhibit high visual quality, with AdvLogo-Hybrid achieving a higher aesthetic score than NAP.}
\label{fig:patch}
\end{figure}

Since our primary goal is to preserve the semantic information of adversarial patches, the visual quality of these patches is a critical consideration. In Fig.~\ref{fig:patch}, we show the patches generated via different attack methods using YOLO v2, YOLO v3, YOLO v4, Faster R-CNN, and SSD. Intuitively, it is apparent that AdvLogo-Hybrid is at least as visually appealing as NAP. To confirm this, we utilize a pretrained aesthetic model based on ViT-L/14 to calculate the aesthetic scores. AdvLogo-Hybrid consistently achieves higher aesthetic scores than its counterpart, NAP. These results, encompassing both attack performance and visual quality, clearly demonstrate the superiority of AdvLogo over NAP.

\subsection{Frequency Domain vs Spatial Domain}

Fig.~\ref{fig:patch} can demonstrate the necessecity of optimization in the frequency domain. When comparing the image quality of logos optimized in the spatial domain versus the frequency domain, distinct differences emerge. Optimization in the spatial domain tends to result in a noticeable loss of visual quality, leading to logos that may be less recognizable or aesthetically pleasing. On the other hand, optimization in the frequency domain effectively preserves the semantic information, maintaining the integrity and clarity of the image. This highlights the advantage of frequency domain optimization in retaining the original appearance and details of the patch while still fulfilling its adversarial purpose.

\subsection{Unconditional Embeddings Optimization}

\begin{table*}
\centering
\caption{Comparison of attack performance of different optimization strategies. The logos are crafted based on YOLO v2.}
\scalebox{0.82}{
\begin{tabular}{c|c c c c c c c|c}
\hline
\textbf{Optimization} & YOLO v2 & YOLO v3 & YOLO v4 & YOLO v4tiny & YOLO v5 & Faster R-CNN & SSD & \textbf{Black-Box Avg$\downarrow$} \\
\hline
Null & 67.52 & 73.01 & 75.89 & 78.13 & 81.99 & 63.31 & 71.18 & 73.92 \\
Embedding & 61.40 & 70.44 & 73.79 & 74.58 & 80.66 & 61.60 & 69.10 & 71.70 \\
Latent & 39.82 & 54.39 & 55.17 & 48.54 & 66.18 & 49.03 & 59.19 & 55.42 \\
Latent$+$Embedding & 37.60 & 48.46 & 45.05 & 33.30 & 52.65 & 37.57 & 47.77 & \textbf{44.13} \\
\hline
\end{tabular}}
\label{tab:optimization}
\end{table*}

To verify the effectiveness of optimizing unconditional embeddings, we conduct comparative experiments with different optimization strategies. As shown in Table~\ref{tab:optimization}, optimizing only the unconditional embeddings slightly improves the attack performance of the logos. However, when both the unconditional embeddings and the latent's frequency domain are optimized together, the attack performance is significantly enhanced compared to optimizing either component alone. These results suggest that optimizing unconditional embeddings can further strengthen adversarial capability. As illustrated in Fig.~\ref{fig:patch}, AdvLogo-Hybrid and AdvLogo-Frequency exhibit significantly different characteristics. This observation suggests that the unconditional embeddings play a crucial role in determining the subarea to which the denoising process converges, highlighting the importance of unconditional embeddings in generating adversarial logos.

\subsection{Robustness to Existing Defenses}
\label{subsec:robustness}

\begin{table*}
\centering
\caption{For AdvPatch and AdvLogo trained on each model, we calculate the average black-box mAP after applying the Segment and Complete (SAC). We use the change in mAP, denoted as $\Delta$, to assess the robustness of the patches.}
\scalebox{0.75}{
\begin{tabular}{c|c|c c c c c c c|c c}
\hline
\textbf{Model} &
\textbf{Method} & YOLO v2 & YOLO v3 & YOLO v4 & YOLO v4tiny & YOLO v5 & Faster R-CNN & SSD & \textbf{Black-Box Avg$\downarrow$} & \textbf{$\Delta\downarrow$}\\
\hline
\multirow{2}{*}{YOLO v2}& AdvPatch & 12.70 & 45.11 & 51.01 & 35.21 & 50.30 & 41.95 & 44.77 & 44.73 & 5.21 \\ 
&AdvLogo(Ours) & 38.02 & 48.85 & 46.56 & 34.10 & 52.98 & 37.80 & 47.97 & \textbf{44.71} & \textbf{0.58} \\
\hline
\multirow{2}{*}{YOLO v3}& AdvPatch & 52.85 & 31.74 & 61.98 & 69.44 & 72.99 & 56.11 & 64.44 & 62.97 & 4.78 \\ 
&AdvLogo(Ours) & 37.49 & 43.02 & 46.16 & 43.00 & 58.22 & 43.16 & 52.74 & \textbf{46.80} & \textbf{-0.12} \\
\hline
\multirow{2}{*}{YOLO v4}& AdvPatch & 52.76 & 56.06 & 30.82 & 40.00 & 61.01 & 51.20 & 51.30 & 52.06 & 12.79 \\ 
&AdvLogo(Ours) &43.48 & 51.92 & 32.95 & 46.03 & 56.18 & 43.08 & 56.84 & \textbf{49.59} & \textbf{0.07} \\
\hline
\multirow{2}{*}{YOLO v4tiny}& AdvPatch & 57.02 & 64.34 & 72.91 & 27.04 & 74.39 & 67.16 & 64.20 & 66.67 & 7.64\\ 
&AdvLogo(Ours) & 43.04 & 48.67 & 46.06 & 23.02 & 41.77 & 42.97 & 51.35 & \textbf{45.64} & \textbf{1.12} \\
\hline
\multirow{2}{*}{YOLO v5}& AdvPatch & 50.32 & 59.20 & 62.93 & 50.73 & 16.06 & 50.38 & 52.80 & 54.39 & 2.96 \\ 
&AdvLogo(Ours) &45.79 & 39.65 & 41.41 & 31.71 & 18.25 & 41.53 & 44.48 & \textbf{40.76} & \textbf{0.00} \\
\hline
\multirow{2}{*}{Faster-RCNN}& AdvPatch & 46.20 & 40.84 & 45.20 & 50.13 & 52.67 & 20.31 & 53.84 & \textbf{48.15} & 19.37 \\ 
&AdvLogo(Ours) & 42.96 & 60.70 & 61.90 & 54.34 & 61.30 & 39.66 & 55.55	& 56.13 & \textbf{1.20} \\
\hline
\multirow{2}{*}{SSD}& AdvPatch & 38.53 & 68.81 & 74.32 & 55.59 & 58.22 & 54.45 & 19.43 & 58.32 & 6.78 \\ 
&AdvLogo(Ours) & 43.22 & 58.92 & 66.35 & 51.76 & 63.01 & 44.19 & 48.00 & \textbf{54.58} & \textbf{0.15} \\
\hline
\end{tabular}}
\label{tab:robustness}
\end{table*}

Segment and Complete (SAC), as proposed by \cite{liu2022segment}, is an effective defense method against adversarial patches. We conduct comparative experiments between AdvPatch and AdvLogo. As shown in Table~\ref{tab:robustness}, AdvLogo outperforms AdvPatch in evading SAC. Specifically, the black-box average mAP of AdvLogo increases by a maximum of 1.20\%, while that of AdvPatch increases by a maximum of 19.37\%. AdvPatch's failure to evade SAC results in AdvLogo surpassing it when defenses are applied, even though AdvPatch may have higher attack performance without defenses for YOLO v2, YOLO v4 and SSD. This suggests that AdvLogo demonstrates stronger robustness to defenses.

\subsection{Comparison of Adversarial Capabilities in Various Semantic Spaces}

\begin{table*}
\centering
\caption{We compare the attack performance across various semantic spaces by varying the objects and tokens. The prompts used for generating adversarial patches are formatted as "a [noun], 8k" and "a [noun], portrait," where the noun represents different objects. In this study, we select "dog" and "tree" as the objects for evaluation. YOLO v2 serves as the white-box model, while the other models function as black-box models.}
\scalebox{0.82}{
\begin{tabular}{c|c c c c c c c|c}
\hline
\textbf{Prompt} & YOLO v2 & YOLO v3 & YOLO v4 & YOLO v4tiny & YOLO v5 & Faster R-CNN & SSD & \textbf{Black-Box Avg$\downarrow$} \\
\hline
a dog, 8k &	37.60 & 48.46 & 45.05 & 33.30 & 52.65 & 37.57 & 47.77 & 44.13 \\ 
a dog, portrait & 34.54	& 47.25 & 43.78 & 45.04 & 55.39 & 49.44 & 53.03 & 48.99 \\
a tree, 8k & 31.21 & 54.87 & 62.15 & 40.40 & 64.45 & 51.17 & 50.98 & 54.00 \\
a tree, portrait & 40.50 & 55.75 & 49.41 & 42.80 & 64.92 & 46.92 & 55.15 & 52.49 \\
\hline
\end{tabular}}
\label{tab:prompts}
\end{table*}

We have thoroughly investigated the adversarial effectiveness using the prompt "a dog, 8k." The prompt is primarily composed by "dog" which is the object and "8k" which is the positive token to guarantee the semantic information. To further validate our hypothesis that every semantic space contains an adversarial subspace, we extend our experiments to various prompts. The attack performance of AdvLogo generated based on YOLO v2 in these semantic spaces is documented in Table~\ref{tab:prompts}. Generally speaking, the attack performance of these patches significantly exceeds that of NAP, which can support our hypothesis. Interestingly, when comparing adversarial effectiveness across different semantic spaces, we find that "dog" exhibits higher attack performance than "tree". Additionally, positive tokens also influence the capability. This may be attributed to the fact that the conditional embeddings in the denoising process, which is fixed in our framework, also have a great impact in adversarial attack. So we can further assume that certain semantic spaces may possess inherently higher adversarial capabilities. However, this assumption warrants further study to fully understand the underlying reasons for these differences in future research.

\section{Future Work}
\label{sec:future work}
Our work has explored the concept of adversarial subspaces and validated that images within these subspaces can be obtained through the diffusion denoising process. In our comparative experiments across different prompts, we observe a distinct variation in attack performance, which may indicate significant differences between various semantic spaces. This observation suggests that different semantic spaces may possess unique adversarial characteristics.

Given these findings, our future work will focus on verifying this assumption. If confirmed, we will explore alternative methods for semantic perturbation by tuning prompts. Additionally, this phenomenon highlights the need for further investigation into the distribution of high-dimensional data, which could provide deeper insights into the behavior of adversarial subspaces.

\section{Conclusion}
\label{sec:conclusion}
In this paper, we propose a novel framework for generating adversarial patches against object detectors. Based on the hypothesis that every semantic space contains an adversarial subspace, we introduce a method for accessing images within these subspaces by modifying the diffusion denoising process. To minimize distribution shifts that could degrade image quality, we apply perturbations in frequency domain. Additionally, we discover that optimizing unconditional embeddings further enhances attack performance. Extensive experiments have validated our hypothesis and demonstrated the effectiveness of our framework.

 {\small
\bibliographystyle{IEEEtranN}

\bibliography{main}
}
 
\vfill

\end{document}